\documentclass[journal,twoside,web]{ieeecolor}
\usepackage{tmi}
\usepackage{cite}
\usepackage{amsmath,amssymb,amsfonts}
\usepackage{algorithmic}
\usepackage{graphicx}
\usepackage{textcomp}

\usepackage{epsfig}
\usepackage{graphicx}
\usepackage{amsmath}
\usepackage{amssymb}

\usepackage{url}
\usepackage{physics}
\usepackage{makecell}
\usepackage{bbm}
\usepackage[font=small,skip=0pt]{caption}
\usepackage{subcaption}
\usepackage{booktabs} 
\usepackage{color, colortbl}
\usepackage{multirow}
\usepackage[mathscr]{eucal}
\usepackage{pifont}% http://ctan.org/pkg/pifont
\DeclareMathOperator*{\argmax}{argmax}

\def\BibTeX{{\rm B\kern-.05em{\sc i\kern-.025em b}\kern-.08em
    T\kern-.1667em\lower.7ex\hbox{E}\kern-.125emX}}
\begin{document}
\title{DSNet: A Dual-Stream Framework for Weakly-Supervised Gigapixel Pathology Image Analysis}
\author{Tiange Xiang, Yang Song, \IEEEmembership{Member, IEEE}, Chaoyi Zhang, Dongnan Liu, \IEEEmembership{Member, IEEE}, Mei Chen, \IEEEmembership{Senior Member, IEEE}, Fan Zhang, Heng Huang, \IEEEmembership{Member, IEEE}, Lauren O’Donnell, Weidong Cai, \IEEEmembership{Member, IEEE}

\thanks{Manuscript received September 4, 2021; revised March 03, 2022; accepted March 06, 2022. (Corresponding author: Weidong Cai.)}
\thanks{Tiange Xiang, Chaoyi Zhang, Dongnan Liu, and Weidong Cai are with the School of Computer Science, University of Sydney, Sydney, NSW 2008, Australia (e-mail: txia7609@uni.sydney.edu.au; tom.cai@sydney.edu.au).}
\thanks{Yang Song is with the School of Computer Science and Engineering, University of New South Wales, Kensington, NSW 2052, Australia (e-mail: yang.song1@unsw.edu.au).}
\thanks{Mei Chen is with Microsoft Corporation, Redmond, WA 98052 USA (e-mail: may4mc@gmail.com).}
\thanks{Heng Huang is with the Department of Electrical and Computer Engineering, University of Pittsburgh, Pittsburgh, PA 15261 USA (e-mail: henghuanghh@gmail.com).}
\thanks{Fan Zhang and Lauren O’Donnell are with the Brigham and Women’s Hospital, Harvard Medical School, Boston, MA 02115 USA (e-mail: fzhang@bwh.harvard.edu; odonnell@bwh.harvard.edu).}
}

\maketitle

\begin{abstract}
	We present a novel weakly-supervised framework for classifying whole slide images (WSIs). WSIs, due to their gigapixel resolution, are commonly processed by patch-wise classification with patch-level labels. However, patch-level labels require precise annotations, which is expensive and usually unavailable on clinical data. With image-level labels only, patch-wise classification would be sub-optimal due to inconsistency between the patch appearance and image-level label. To address this issue, we posit that WSI analysis can be effectively conducted by integrating information at both high magnification (local) and low magnification (regional) levels. We auto-encode the visual signals in each patch into a latent embedding vector representing local information, and down-sample the raw WSI to hardware-acceptable thumbnails representing regional information. The WSI label is then predicted with a Dual-Stream Network (DSNet), which takes the transformed local patch embeddings and multi-scale thumbnail images as inputs and can be trained by the image-level label only. Experiments conducted on three large-scale public datasets demonstrate that our method outperforms all recent state-of-the-art weakly-supervised WSI classification methods.
\end{abstract}


\section{Introduction}
\label{introduction}
\IEEEPARstart{C}{onvolutional} Neural Networks (CNNs) have shown their extraordinary capabilities in feature extraction in image analysis tasks. With the help of CNNs, great progress has been made in pathology image analysis \cite{tokunaga2019adaptive, li2019attention, maksoud2020sos, xiang2020bio,song2017low,wang2021bix}. However, whole slide images (WSIs) cannot be analyzed the same way as typical digital images due to their gigapixel resolution. To achieve a feasible and effective WSI analysis (e.g. classification), most of the existing methods adopt a patch-based approach that first divides the WSI into small patches and then predicts the WSI-level label based on the predicted labels of individual patches \cite{wang2019rmdl, wang2016deep, bejnordi2017diagnostic,araujo2017classification, liu2020pdam, zheng2018histopathological, mercan2017multi}. Unfortunately, such fully-supervised approach relies on patch-level ground truth annotations, which require time consuming efforts from pathology experts. Meanwhile, image-level labels of WSIs are more readily attainable, therefore we seek to answer the question of whether a weakly supervised approach employing image level WSI labels can achieve comparable performance.

%Therefore, classifying WSIs in a weakly-supervision style by using their image-level labels only is of great significance. 

\begin{figure}
		\begin{center}
			\includegraphics[width=1\linewidth]{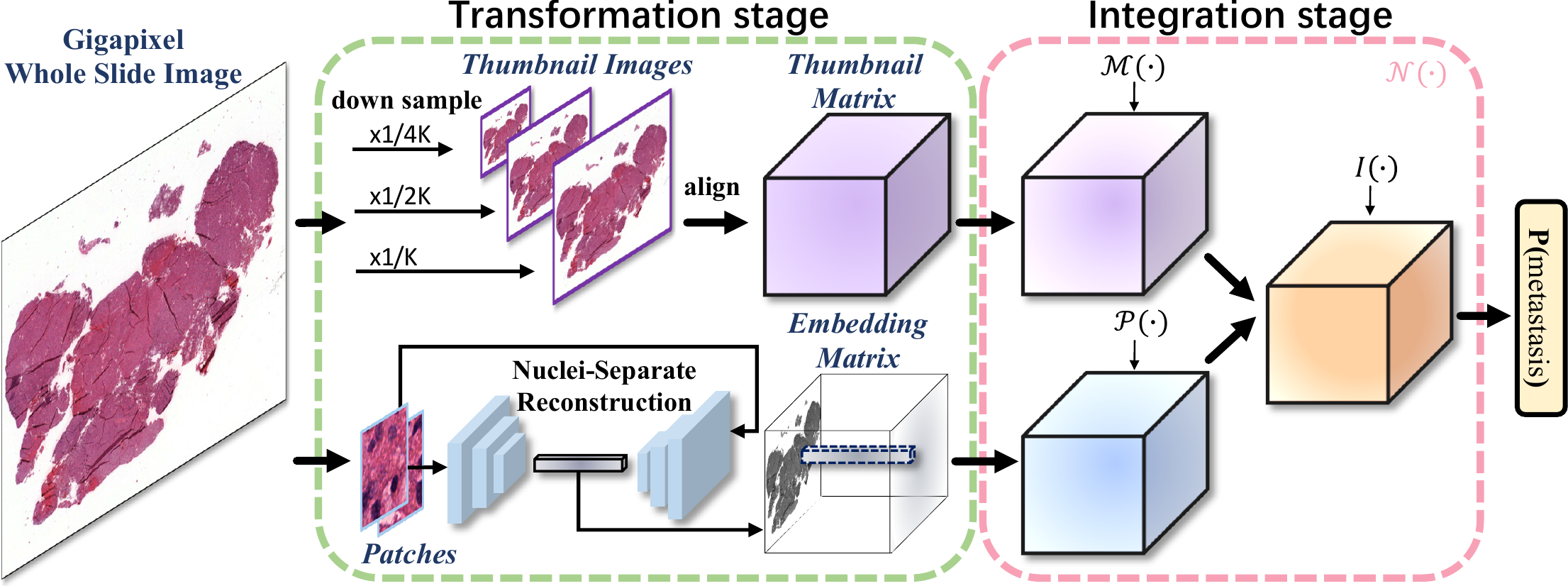}
		\end{center}
		\caption{Our WSI classification framework consists of a \textbf{Transformation} stage followed by an \textbf{Integration} stage. Our \textbf{Integration} stage is achieved by the proposed DSNet $\mathcal{N}(\cdot)$, which is comprised of different building blocks $\mathcal{M}(\cdot)$ and $\mathcal{P}(\cdot)$ for each stream, and ends up with an adaptive aggregation $\mathcal{I}(\cdot)$ of the two streams.}
		\label{fig1}
	\end{figure}
% 	\cite{hou2016patch,zhang2018whole,DBLP:journals/corr/abs-1802-02212,chen2019rectified,spanhol2016breast, lu2020data} 

	%Time spent on annotating WSI patches can be saved, allowing expedient therapies and treatments to be provided in time. Also, an efficient detection and classification of cancer metastasis assists the research and development of targeted medicines.
	
% 	Compared to a fully-supervised approach that uses large number of patch-wise annotations as extra supervision signals, we focus on a weakly-supervised approach that only
	
	We are mindful to ensure that our weakly-supervised approach is able to fit into real clinical workflow, which pathologists typically analyze WSIs by first skimming through a thumbnail of the WSI to look for possible regions of interest that may contain cancerous tissues. Then, they zoom in on the regions of interest to investigate the local details. While the above process resembles the attention mechanism \cite{wang2017residual, vaswani2017attention}, such a mechanism cannot be applied directly to WSIs due to their gigabyte sizes. We propose a whole image-based approach where gigapixel WSIs are first transformed into two matrices in compact latent spaces representing local and regional descriptors of the raw WSI. Then, a Dual-Stream Network (DSNet) integrates the compressed representations through a stack of Concurrent Bottleneck blocks. The two matrices are adaptively aggregated by assigning importance scores through both stream-wise and channel-wise attentions. Intuitively, our DSNet uses the coarse visual clues implied in thumbnails to guide detailed analysis at zoomed-in regions of interest. Our weakly-supervised framework is outlined in Figure \ref{fig1}.

	\textcolor{black}{The overall technical novelty of our work is three-fold: \textbf{(i)} We proposed a global-local framework that achieves weakly supervised WSI analysis based on multi-scale thumbnails and neural encoded embeddings, which differs from \cite{tellez2019neural} that classifies WSI embedding directly and \cite{chen2019GLNET} that uses raw RGB images (and patches) only. \textbf{(ii)} We improved the unsupervised WSI encoding schema \cite{tellez2019neural} by adapting S-CAE \cite{hou2019sparse} with non-trivial upgrades to achieve nuclei-tissue separate encoding to sift out non-informative visual redundancies. \textbf{(iii)} We optimized the network architecture considering the sparsity issues in WSI analysis, and proposed multiple novel operators and modules for better effectiveness and efficiency.  We validated our method on three large-scale benchmark datasets Camelyon16 \cite{bejnordi2017diagnostic}, TCGA-LUSC \cite{tcga}, and BCNB \cite{bcnb}, and our method outperforms recent state-of-the-art weakly-supervised methods with significantly fewer network parameters.}

	Our framework can be easily extended to other WSI-related tasks, such as thumbnail-wise tumor region segmentation and the metastasis localization. With abundant local and regional information provided by our framework, our DSNet can be easily modified to an encoder-decoder structure to fit different supervision signals. 
	
	\section{Related Work} \label{relatedwork}
	
	\subsection{Weakly-supervised WSI Classification} 
	Classifying images that are too large to fit into the computer memory is a challenging topic \cite{tokunaga2019adaptive, maksoud2020sos}. State-of-the-art classification methods on extremely high resolution images, such as satellite images \cite{sankaran2015low, yuan2015survey} and WSIs \cite{bejnordi2017diagnostic,wang2016deep, wang2019rmdl, wang2018weakly,araujo2017classification, lin2019fast, hosseini2019focus} require fully annotated patch-level annotations as a guarantee, which are expensive to acquire and unavailable for most datasets.
	
	Without patch-wise annotations, WSI classification can be approached in a weakly-supervised manner with image-wise \cite{tellez2019neural} or patch-wise \cite{chikontwe2020multiple, vu2020novel} designs. As a representative image-wise approach, Tellez \textit{et al.} \cite{tellez2019neural} used a Generative Adversarial Network \cite{goodfellow2014generative} to compress WSI into the latent space with a smaller size, and then employed a standard deep learning model to make predictions.
	
	Differing from image-wise methods, patch-wise methods followed a Multi Instance Learning (MIL) paradigm, where groups of instances (bags) are analyzed based on the group label. When determining the presence of tumor or metastasis in WSIs, positive bags indicate WSIs with at least one patch containing diseases and negative bags when all patches in the WSIs are classified negative. Chen \textit{et al.} \cite{chen2019rectified} designed a rectified cross-entropy loss to model cancer metastasis patterns. Hou \textit{et al.} \cite{hou2016patch} proposed the usage of a fusion model that sifts out discriminative patches by utilizing a patch level CNN. Instead of selecting discriminative patches only, Chikontwe \textit{et al.} \cite{chikontwe2020multiple} designed a center loss that maps patch embeddings to a single image embedding centroid and reduced the intra-class variations. \cite{hashimoto2020multi} proposed a similar multi-scale patch-wise approach for analysing WSIs with the help of domain adversarial learning. With the help of GANs, \cite{zhao2020predicting} proposed a graph convolutional network for learning bag-level representations. However, the proposed VAE-GAN relies on extra domain knowledge and annotations from the WSI experts. In a recent work, \cite{lu2021ai} designed a multi-task classification framework for predicting the source of metastasis in WSI slides with low resource costs. However, most of the patch-wise weakly-supervised methods have some evident drawbacks: localizing discriminative patches can be difficult and inaccurate due to their restricted receptive field. Also, the patch-wise processing requires inference of all patches, which is time consuming. \textcolor{black}{Although the workflows of DSNet and GLNet \cite{chen2019GLNET} share a similar design, our framework differs from GLNet in four aspects. \textit{(i)} Both branches in GLNet take the raw RGB images (in different resolutions) as inputs, while DSNet only accepts down-sampled thumbnail images in the global branch. In the local branch, DSNet is fed with the abstract embedding matrix that sifts out the visual redundancies and contains only meaningful semantics, which fosters the optimization of DSNet further. \textit{(ii)} We considered the sparsity issue in WSIs and designed multiple sparsity-aware operators to be incorporated in the DSNet. \textit{(iii)} Multiple novel operators (multi-scale convolution block, concurrent bottleneck and adaptive aggregation) were designed that prove to be effective. \textit{(iv)} Our framework was designed in a resource-aware style with both minimum network parameters and inference latency, while GLNet focuses on segmentation accuracy only.}

% 	Differing from GLNet \cite{chen2019GLNET} that also adopts a global-local framework under a feature sharing design, we employ local path encoding to initially sift out overwhelming visual redundancies, while the multi-scale thumbnail images and patch embeddings could then be independently processed through our two-branch network. Moreover, we engineer and develop an efficient and effective DSNet with simple building blocks to better capture the global-local underlying patterns.

	 %\cite{tellez2019neural} introduces a similar encoding framework to ours, however, we encode nuclei separately from background tissues towards better semantic representation.
	%Instead of using patches only, our method utilizes information at both local (patch) and regional (thumbnail) levels and classifies the encoded matrices effectively and efficiently.

% 	The performance of MIL-based methods is mainly limited at its restricted receptive field. Although higher-level information can be gathered with multi-scale learning techniques \cite{hashimoto2020multi}, representative regional features are still hard to acquire unlike image-wise approaches do. 
	
	%Zhang \textit{et al.}. \cite{zhang2018whole} sifted patches on the basis of the presence of different cancer metastasises. 
	\subsection{Attention Mechanisms} 
	The attention mechanism can be interpreted as enhancing informative signals while suppressing useless ones \cite{itti2001computational, itti1998model,olshausen1993neurobiological,chen2020computer}. It has been applied to a wide variety of visual applications recently \cite{Chen_2020_CVPR, Zoran_2020_CVPR, hu2020span, mejjati2020look, zhang2020resnest}. For example, Wang \textit{et al.} \cite{wang2017residual} used extra encoder-decoder style branches to model the spatial attention at multiple levels in a CNN. The proposed network significantly boosts performance in image classification by denoising the intermediate feature maps. Hu \textit{et al.} \cite{hu2018squeeze} investigated attention from channel-wise perspective. The proposed SENet self-recalibrates different feature maps regardless of their spatial signals. Furthermore, Roy \textit{et al.} \cite{roy2018concurrent} and Woo \textit{et al.} \cite{woo2018cbam} introduced novel attention strategies by combining the spatial-wise and channel-wise attentions to further enhance the ability of the CNNs. Following the above methods, GSoP \cite{gao2019global} designed a second-order pooling operator, which utilizes the covariance matrix for learning the channel-wise attentions. SKNet \cite{li2019selective} proposed to adjust receptive field size of convolution kernels dynamically based on multi-scale features. In a recent work, ECA-Net \cite{wang2020eca} proposed a more efficient channel attention framework, that attention logits can be learned with a constant number of parameters. Unlike the above methods, we apply concurrent stream-wise and channel-wise attentions for an adaptive aggregation of features encoded at different magnifications. 

	\section{Method} \label{method}
	
	\subsection{Framework} \label{WAC}

	%The biggest challenge lie in the designation of an efficient mapping algorithm from raw visual signals in WSI to a compact representation, while maintaining as much information as possible. 
	As outlined in Figure \ref{fig1}, we define our image-level weakly-supervised WSI classification framework as a two-stage sequential process: an efficient \textbf{Transformation} from gigapixel WSIs into a compact latent space, followed by an effective \textbf{Integration} that captures and fuses the underlying patterns encoded in the transformed representations.
	
	During the \textbf{Transformation} stage, our framework takes the gigapixel WSI $\mathbf{X}$ as input and converts it into two correlated compact representations, which are the \textit{thumbnail matrix} $\mathbf{T}$ and the \textit{embedding matrix} $\mathbf{V}$, such that:
	\begin{equation} \label{eq1}
	||\mathbf{T}||\ +\ ||\mathbf{V}||\ \ll ||\mathbf{X}||,
% 	\mathbb{R}^\mathbf{(V)}\ +\ \mathbb{R}^\mathbf{(T)}\ \ll \mathbb{R}^\mathbf{(X)}
	\end{equation}
	where $||\cdot||$ denotes the number of pixels. Given the constraint of Eq. \ref{eq1}, we adopt an unsupervised transformation strategy to ensure the latent representations keep as much discriminative information in $\mathbf{X}$ as possible. Subsequent to the \textbf{Transformation} stage, our \textbf{Integration} process (Sec. \ref{DCR}) can be formulated as:
	
% 	\begin{equation}
% 	\mathbf{Pr}\left(\mathbf{X},y\right)\ \to\ \argmax_{\theta} \mathbf{Pr} \big(\mathcal{N}\left(\mathbf{T},\mathbf{V};\theta\right),y\big),
% 	\end{equation}
	
	\begin{equation}
	\argmax_{\theta} \mathbf{Pr} \big(\mathcal{N}\left(\mathbf{T},\mathbf{V};\theta\right),y\big),
	\end{equation}
	where $y$ is the WSI label, $\mathcal{N}(\cdot)$ and $\theta$ are our DSNet and its model parameters, respectively.
	
	The \textit{thumbnail matrix} $\mathbf{T}$ of $\mathbf{X}$ is comprised of thumbnail images at different magnification levels. Practically, thumbnails of different levels can be obtained by downsampling $\mathbf{X}$ using a set of pre-defined factors $\{\text{K}\}$ to hardware-acceptable sizes. We then construct $\mathbf{T}$ by nearest neighbor interpolation of the smaller thumbnails to align with the finest-scale image. The resultant matrix is regarded as a global representation of $\mathbf{X}$ that encodes coarse-scale multi-level regional information.

	\subsection{Patch Encoding} \label{encode}
	
	Sharing the same spatial correspondences as the \textit{thumbnail matrix} $\mathbf{T}$, the \textit{embedding matrix} $\mathbf{V}=\{\mathbf{v}_{i,j}\}$ of WSI $\mathbf{X}=\{\mathbf{x}_{i,j}\}$ encodes local information at the patch level. We assume that not all visual information in a patch $\mathbf{x}_{i,j}\in \mathbb{R}^{\text{S}\times \text{S}\times 3}$ contributes to cancer diagnosis, thus $\mathbf{V}$ is constructed by removing visual redundancies in every patch of $\mathbf{X}$. We denote the representative vector $\mathbf{v}_{i,j}\in \mathbb{R}^{1\times 1\times \text{C}}$ as the compressed patch $\mathbf{x}_{i,j}$. An autoencoder network is adopted to formulate the mapping between $\mathbf{v}_{i,j}$ and $\mathbf{x}_{i,j}$ in an unsupervised style by reconstructing $\mathbf{x}_{i,j}$. The autoencoder is then slid throughout all spatial locations $i,j$ with a stride of $\text{S}$ in both directions to generate a complete set of non-overlapping representation vectors $\mathbf{v}_{i,j}$ that eventually construct the \textit{embedding matrix} $\mathbf{V}$. The above process maps a patch in the input WSI to a pixel in $\mathbf{V}$ with visual redundancies removed.
	
\begin{figure}[t]
		\begin{center}
			\includegraphics[width=1.0\linewidth]{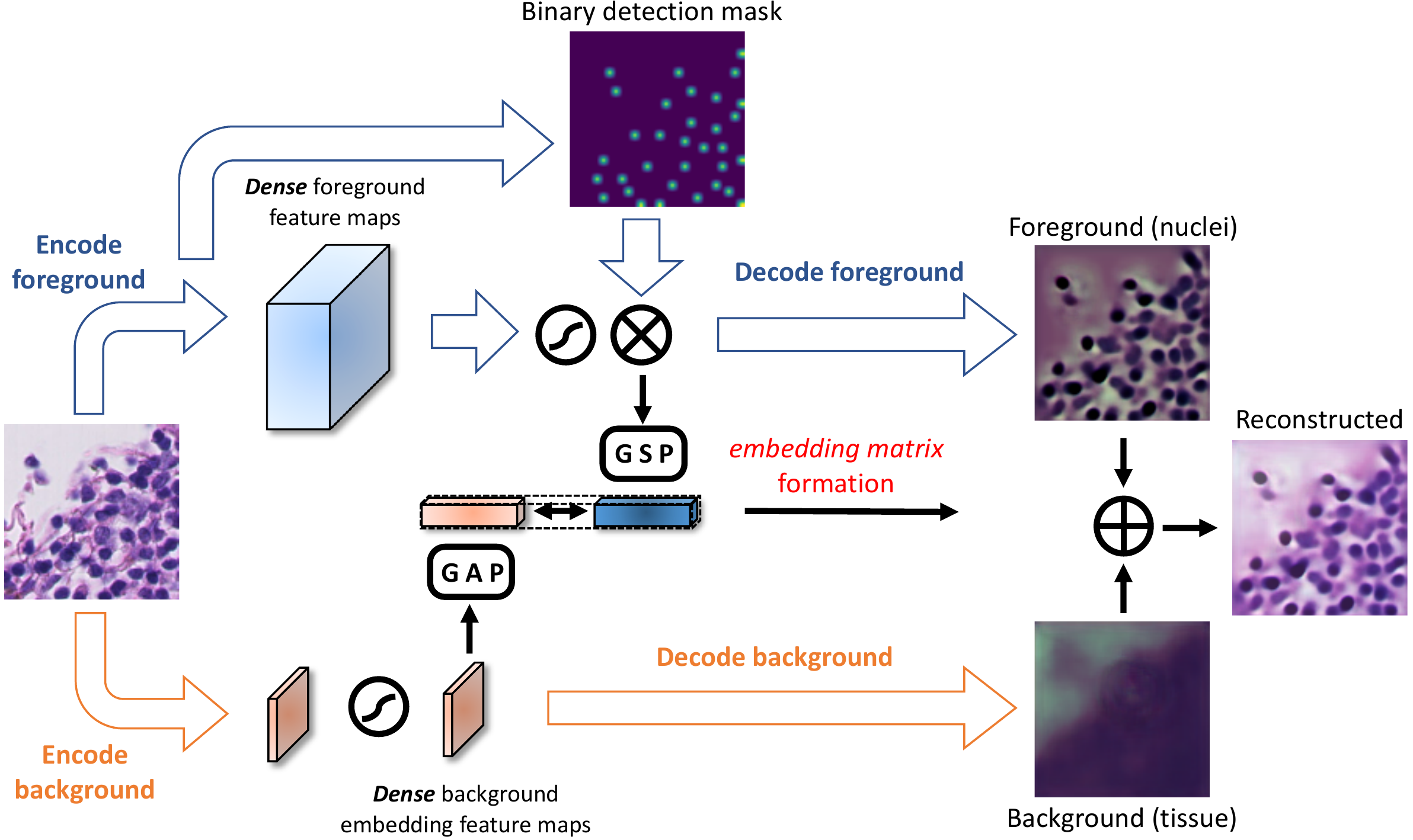}
		\end{center}
		\caption{\textbf{Our embedding matrix encoding process.} Representative vector of input patch is constructed as the concatenation of the pooling results of sparse foreground embeddings and dense background embeddings.}
		\label{fig:encode}
	\end{figure}

	Our patch encoding process is outlined in Figure \ref{fig:encode}. It follows an autoencoder structure based on the \textbf{S}parse \textbf{C}onvolutional \textbf{A}uto\textbf{E}ncoder (S-CAE) \cite{hou2019sparse} with two branches to encode the foreground and background embeddings separately. The encodings are then separately decoded and summed up together to reconstruct the input patch. Differing from the mixed encoding in conventional autoencoders, we distinguish the nuclei from tissues to achieve a foreground-background separate encoding. 
	
	We encode the background features as small dense embedding feature maps $\mathbf{B}$. The majority of the patch background is tissue with relatively homogeneous color and texture, which does not carry sufficient diagnostic information. Therefore, a set of compact feature maps is adequate to describe the background tissue patterns. 
	
	\begin{figure*}[t]
		\begin{center}
			%\fbox{\rule{0pt}{2in} \rule{.9\linewidth}{0pt}}
			\includegraphics[width=1.0\linewidth]{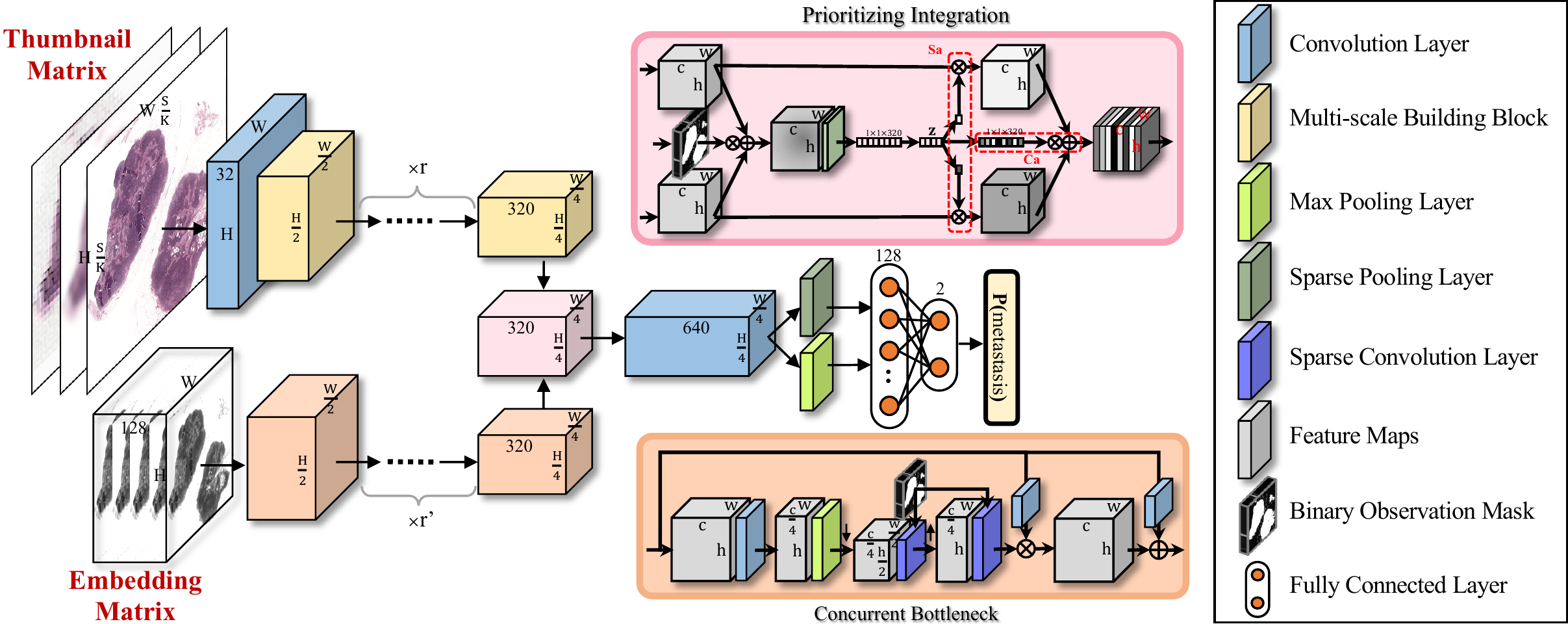}
		\end{center}
		\caption{\textbf{Overview of our DSNet architecture.} Relative sizes of feature maps are shown in this figure. `$\times$r' denotes repeating the same block by r times. \textbf{Sa} denotes the stream-wise attention and \textbf{Ca} the channel-wise attention. More basic operators are shown in Figure \ref{fig:comp}.}
		\label{fig:network}
	\end{figure*}
	
	\begin{figure}[t]
	\begin{center}
		\includegraphics[width=1.0\linewidth]{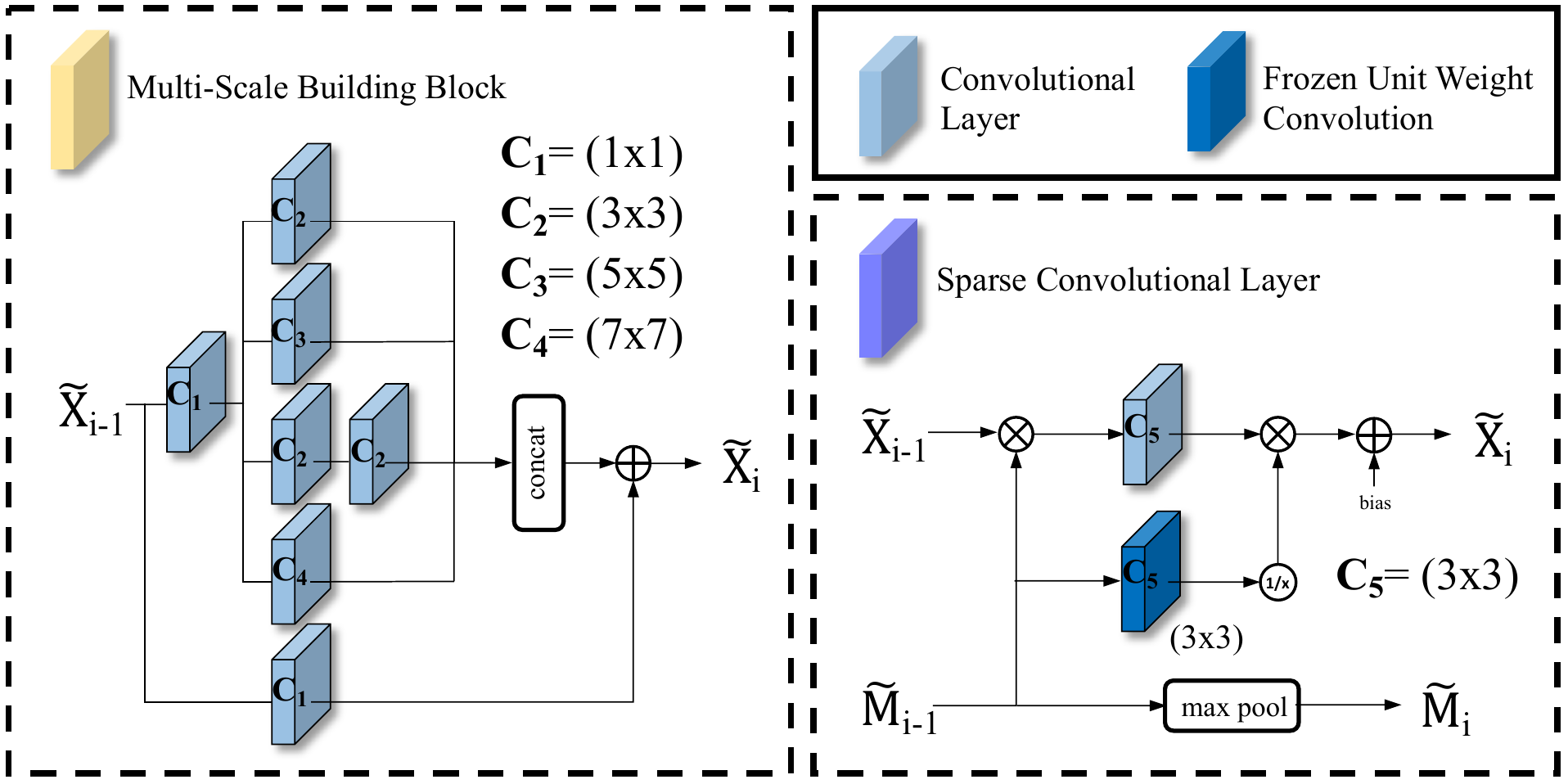}
	\end{center}
	\caption{\textbf{Details of the basic operators used in DSNet}. $\Tilde{\text{X}}$ and  $\Tilde{\text{M}}$ denote feature maps and binary observation mask, respectively. ``concat'' denotes the concatenation operation, ``max pool'' denotes max pooling operation and $1/x$ denotes element-wise inversion. Kernel sizes are illustrated as well.}
	\label{fig:comp}
\end{figure}
	
	Unlike background tissues, the foreground nuclei contain more complex patterns and require more descriptive feature representations. The dispersed nuclei can be encoded by activating the tensor units near the nuclei only. The small amount of nuclei existed in each patch would lead to \textit{sparse} foreground embedding feature maps $\mathbf{F}$. To ensure the crosswise sparsity \cite{hou2019sparse} of the nuclei in any foreground feature map $\mathbf{F}_{l} \in \mathbf{F}$, a binary detection mask $\mathbf{M}$ can be generated. $\mathbf{M}^{i,\ j}$ indicates whether a nucleus appears at location ($i, j$) and the corresponding $\mathbf{F}^{i,\ j}_l$ is activated, such that:
	\begin{equation} \label{binarymask1}
	\mathbf{M}^{i,\ j} = \mathbbm{1}\big(\sum_{l}\mathbbm{1}(\mathbf{F}^{i,\ j}_{l}\neq0) > 0\big),
	\end{equation}
	where $\mathbbm{1}(\cdot)$ is an indication function that returns 1 if the condition is true and 0 otherwise. Note that our binary detection mask $\mathbf{M}$ can be directly computed based on $\mathbf{F}$, and no gradient is required. The gradient flow from decoders to encoders is then ensured by using an extra branch. 
	
	Eq. \ref{binarymask1} yields that, if any feature channel of a pixel in $\mathbf{F}$ carries useful information, we presume a nucleus can be observed at such location, and set the mask value to 1. However, such condition can be hardly satisfied in practice, and an alternative strategy is to pre-set a fixed number of neurons with the smallest values to be activated in $\mathbf{M}$. In reality, the number of nuclei in different patches would vary significantly, and units in $\mathbf{M}$ are better activated adaptively. We therefore set an adaptive sparsity rate $\rho$ to control the activation rates of the nuclei across different patches. With $\rho$, $\mathbf{M}$ is now computed by thresholding $\mathbf{F}$ at $\rho_{th}$ percentile for all spatial locations. The sparsity rate is then updated following the running average approach \cite{Ioffe:2015:BNA:3045118.3045167} with a pre-defined momentum. A high $\rho$ encodes fewer nuclei and a low $\rho$ confuses foreground with more background tissues, hence the optimal $\rho$ can be decided from a binary search performed on each dataset. Finally, we get the sparse foreground feature maps $\mathbf{F}$ as the element-wise multiplication of $\mathbf{M}$ and the densely encoded foreground features.

	 Sigmoid gates are employed to normalize the embedding feature maps by restricting the features within (0, 1). By global averaging on dense background embedding $\mathbf{B}$ and sparse $\mathbf{F}$ (discarding the 0 elements), the representative vector $\mathbf{v}_{i,\ j}$ for the patch $\mathbf{x}_{i,\ j}$ can be obtained as the concatenation of the background and foreground embedding features. Our DSNet $\mathcal{N}$ then takes and fuses \textit{thumbnail matrix} $\mathbf{T}$ and \textit{embedding matrix} $\mathbf{V}$ to produce the final prediction.
	 
	 Compared to the original S-CAE \cite{hou2019sparse}, instead of adopting as a simple nuclei detection tool, the encoded sparse foreground and dense background features are further processed to be better adopted into our DSNet. \textcolor{black}{To this end, we upgrade the original S-CAE architecture along the following dimensions: \textit{(i)} We used more advanced building blocks \cite{he2016deep, han2020ghostnet} to extract low-level features in the image patches. \textit{(ii)} We alleviated the artifacts shown in the decoding process by optimizing the upsampling technique with bilinear resizing. \textit{(iii)} We incorporated a feature bottleneck (Fig. \ref{fig:encode}) by using global average pooling and the proposed global sparse pooling to transfer the task from nuclei detection (the original objective of S-CAE) to our embedding matrix encoding.}
	 
% 	 To this end, we modify the original S-CAE architecture by employing more advanced building blocks to extract low-level features and upgrading the decoder layers to eliminate decoding artifacts.
	
	 Converging patch reconstruction loss proves that there exists at least one function (the learned decoder) that is able to understand and capture the encoded information in $\mathbf{V}$. Therefore, with supervision, our DSNet may be also capable of decoding the underlying patterns implied in $\mathbf{V}$. At the unsupervised \textbf{Transformation} stage, no diagnostic information is expected to be encoded in $\mathbf{V}$, as we delegate the semantic reasoning job to the DSNet during the \textbf{Integration} stage with the supervision of WSI labels. 
	
	\subsection{Dual-Stream Network} \label{DCR}
	Our \textbf{Integration} stage is to take $\mathbf{T}$ and $\mathbf{V}$ of a WSI and infer the WSI-level label through the proposed DSNet. The design challenges of this stage include: \textbf{(1)} how to efficiently capture the encoded patterns in $\mathbf{V}$ and $\mathbf{T}$, \textbf{(2)} how to adaptively aggregate $\mathbf{V}$ and $\mathbf{T}$ towards the final prediction, and \textbf{(3)} how to alleviate the sparsity issue encountered in the representation matrices. To accomplish \textbf{(1)}, we design basic building blocks to extract discriminative features from each representation matrix. To achieve \textbf{(2)}, we employ an attention-based weighting process to obtain the aggregation adaptively. To tackle \textbf{(3)}, we replace the dense operators with the sparsity-aware ones to function on such sparse data.

    \paragraph{Building blocks} The \textit{thumbnail matrix} $\mathbf{T}$ is comprised of raw RGB images at different magnifications. We adopt the commonly used Multi-Scale blocks \cite{li2019attention} to extract varying range regional information in $\mathbf{T}$. For better computational efficiency, we replace the standard convolutions with their separable variants \cite{howard2017mobilenets}. Unlike the highly squeezed thumbnail images, the neural encoded \textbf{embedding matrix} $\mathbf{V}$ implies great spatial-wise redundancies. The intuition is that in WSIs, adjacent patches usually have similar appearances and semantics, hence it would be sufficient to apply convolutions on the local representative of a neighborhood (e.g. the maximum values) instead of all nearby units. To better discriminate local features in $\mathbf{V}$, we employ a max pooling layer to create spatial bottlenecks. A paired upsampling layer is then utilized to map the max pooled tensor back to its original size, and we insert the convolutions between and after the spatial bottleneck. We embed the spatial bottleneck into a channel bottleneck block \cite{he2016deep}, constructing the Concurrent Bottleneck building blocks for $\mathbf{V}$. 
    
    %For notation simplicity, we denote the \textbf{M}ulti-\textbf{S}cale building block as MS, and the \textbf{C}oncurrent \textbf{B}ottleneck building block as CB.
	
	\paragraph{Adaptive aggregation} At the end of our DSNet, the two stream features are eventually aggregated toward the final label prediction. We achieve this in a self-attention manner. First, a sigmoid gated stream-wise attention \cite{li2019selective} score is generated to scale all features in one stream, whose residual of 1 is multiplied to the other stream broadcastly. We then learn a subsequent set of channel-wise attention \cite{hu2018squeeze} to the element-wise summation of the two scaled streams.

% 	\textbf{Sparsity in the embedding matrix.} Usually, more than 50\% of a WSI is covered by connective tissue, which does not provide useful diagnostic information, and can be excluded during pre-processing \footnote{We exclude less informative patches based on their pixel intensities.}. Sifting out the large number of uninformative patches and replacing them with zero introduces significant sparsity in $\mathbf{V}$. Standard dense operators (e.g. pooling, normalization) without ignoring the zero values perform poorly on such sparse WSI representations. To tackle the problem, we apply sparse convolution \cite{uhrig2017sparsity} on the sparse $\mathbf{V}$ for basic feature aggregation. Compared to the conventional convolution operator, an extra binary observation mask is constructed indicating whether a pixel in $\mathbf{V}$ (a patch in WSI) is valid or not. To better handle the sparsity in $\mathbf{V}$, we modify the dense pooling and normalization operators by masking the sparse $\mathbf{V}$ first, and then calculate on the non-zero values only.
    \paragraph{Sparsity in the embedding matrix} Usually, more than 50\% of a WSI is covered by connective tissue, which does not provide useful diagnostic information, and can be excluded during pre-processing (Sec. \ref{pre}). Sifting out the large number of uninformative patches and replacing them with zero introduces significant sparsity in $\mathbf{V}$. Standard dense operators (e.g. pooling, normalization) without ignoring the zero values perform poorly on such sparse WSI representations. To tackle the problem, we apply sparse convolution \cite{uhrig2017sparsity} on the sparse $\mathbf{V}$ for basic feature aggregation. Compared to the conventional convolution operator, an extra binary observation mask is constructed indicating whether a pixel in $\mathbf{V}$ (a patch in WSI) is valid or not. 
    
    To better handle the sparsity in $\mathbf{V}$, we modify the dense pooling and normalization operators by masking the sparse $\mathbf{V}$ first, and then calculate on the non-zero values only. Conventional average calculation operator averages the values across all spatial locations. However, applying such operator directly on a sparse tensor might suffer from a significant information under-weighting, as a large amount of zeros are also taken into the calculation. With the help of the observation mask, we derive the sparse mean by neglecting all zero values and averaging over the masked entries only. We modify the variance calculation operator on sparse data by using the observation mask as well. We compute the sparsity-aware mean and variance as:
    
    \begin{equation} \label{mean}
    \texttt{mean}_{s}(\mathbf{F}) = \frac{\sum_{i,j}(\mathbf{F}_{i,j}*\mathbf{O}_{i,j})}{\sum_{i,j}\mathbf{O}_{i,j}},
    \end{equation}
    \begin{equation} \label{var}
    \texttt{var}_{s}(\mathbf{F}) = \frac{\sum_{i,j}(\mathbf{F}_{i,j}-\texttt{mean}_{s}(\mathbf{F}))^2 - \beta} {\sum_{i,j}\mathbf{O}_{i,j}},
    \end{equation}
    
    where $\mathbf{F}$ is a sparse feature map, $\mathbf{O}$ is the binary observation mask, and overweight $\beta$ is computed as $\sum_{i,j}(1-\mathbf{O}_{i,j})*\texttt{mean}_{s}(\mathbf{F})^2$ indicating the over-weighted mean values brought by the zero entries.

	\paragraph{Network structure} As the name suggests, our DSNet processes the two matrices $\mathbf{T}$ and $\mathbf{V}$ in two separate streams with stacks of the aforementioned multi-scale blocks and concurrent bottleneck blocks. The local and regional feature maps are eventually fed into the adaptive aggregation block to assign importance scores to each of the streams and their integration respectively. Similar to \cite{sandler2018mobilenetv2}, the classification head starts with a point-wise convolution that maps the aggregated features to a higher dimension. A sparse pooling layer together with a max pooling layer are then employed to generate global descriptors by summarizing over spatial dimensions. Our DSNet ends with two fully-connected layers to regress the final classification scores. Figure \ref{fig:network} illustrates an overview of our DSNet architecture, and the building block details are presented in Table \ref{tab:param}.

\begin{table}[t]
		\begin{center}
			\caption{Building block parameter details in our DSNet. MS denotes the Multi-Scale block, CB the Concurrent Bottleneck block.}
			\label{tab:param}
			\resizebox{\linewidth}{!}{\begin{tabular}{c c c c c c}
				\toprule
				%\Xhline{3\arrayrulewidth}
				Stream & Block & \#in & \#out& kernel size & stride \\
				\hline
				\hline
				Thumbnail & Conv & 3 * L & 32 & 7 & 2 \\
				Thumbnail & MS & 32 & 64 & 3,5,7 & 1 \\
				Thumbnail & MS & 64 & 144 & 3,5,7 & 2 \\
				Thumbnail & MS & 144 & 256 & 3,5,7 & 2 \\
				Thumbnail & MS & 256 & 320 & 3,5,7 & 1 \\
				Embedding & CB & 128 & 144 & 5 & 2 \\
				Embedding & CB & 144 & 224 & 5 & 1 \\
				Embedding & CB & 224 & 256 & 5 & 2 \\
				Embedding & CB & 256 & 320 & 5 & 1 \\
				\bottomrule
				%\Xhline{3\arrayrulewidth}
			\end{tabular}}
		\end{center}
		
\end{table}

% 	\begin{table}[t]
% 		\begin{center}
% 			\caption{Component studies on the testing set of Camelyon16. \textbf{SC} represents sparse convolutions, \textbf{MS} the Multi-Scale convolution, \textbf{Pw} the Priority weighting operator, \textbf{Iw} the Integration weighting operator.}
% 			\label{tab:ablation}
% 			\resizebox{\linewidth}{!}{\begin{tabular}{c| c c c c c}
% 					\toprule
% 					Methods & Accuracy & AUC & Precision & Recall & $\text{F}_1$-score \\
% 					\hline
% 					\hline
% 					%\Xhline{2\arrayrulewidth}
% 					w/o \textbf{SC} & 51.3\% & 0.589 & 0.630 & 0.542 & 0.582  \\
% 					w/o thumbnail & 68.0\% & 0.689 & 0.701 & 0.847 & 0.768 \\
% 					w/o \textbf{CB} & - & - & - & - & - \\
% 					w/o \textbf{MS} block & 64.3\% & 0.627 & 0.701 & 0.750 & 0.725\\
% 					w/o \textbf{DA} & 63.5\% & 0.654 & 0.634 & \textbf{0.986} & 0.772  \\
% 					w/o \textbf{Pw} w/ \textbf{Iw} & 56.0\% & 0.670 & 0.700 & 0.514 & 0.592  \\
% 					w/ \textbf{Pw} w/o \textbf{Iw} & 60.9\% & 0.714 & 0.702 & 0.653 & 0.676 \\
% 					\textbf{Full DSNet} & - & - & - & - & - \\
					
% 					\bottomrule
% 					%\Xhline{3\arrayrulewidth}
% 			\end{tabular}}
% 		\end{center}
% 	\end{table}

	\section{Experiments} \label{exp}
	In this section, empirical experiments are performed to validate the effectiveness of the proposed weakly-supervised image-level method in WSI classification task. 
	
	\subsection{Dataset Description and Preparation}
	 In this work, our method was evaluated on three large-scale public available datasets: Camelyon16 \cite{bejnordi2017diagnostic}, The Cancer Genome Atlas Lung Squamous Cell Carcinoma project (TCGA-LUSC) \cite{tcga}, and Early Breast Cancer Core-Needle Biopsy WSI (BCNB) \cite{bcnb}. 
	
	\paragraph{Camelyon16}
	The Camelyon16 dataset contains 400 sentinel lymph node Hematoxylin and Eosin (H\&E) stained WSIs from breast cancer patients. We followed the official dataset split with 270 training WSIs and 130 validation WSIs. 110 of the 270 training WSIs are positive cases (metastasis) and the rest 160 are negative cases (normal). The 130 validation WSIs comprised of 50 positive cases and 80 negative cases.
	
	To construct the embedding matrix $\mathbf{V}$, raw WSIs are segmented into $256\times 256$ patches to be fed into our S-CAE. There is an average number of 140296 patches in the Camelyon16 WSIs. \textcolor{black}{However, training S-CAE with all of the data is time consuming and unnecessary}, we randomly drawn 600 patches from each of the 20 randomly selected WSIs in the positive training set, negative training set, and validation set respectively to conduct network training. In order to better encode the patches with a rich distribution of nuclei, we ensure at least 90\% of the training patches are drawn with all three channels' intensities being less than 70\% of the maximum intensity, which represents samples from rich nucleus regions.
	
	\paragraph{TCGA-LUSC}
	The Cancer Genome Atlas (TCGA) is a public database that collects H\&E stained WSIs across various cancers and patients. We evaluated our method on 500 WSIs randomly sampled from the Lung Squamous Cell Carcinoma (LUSC) project for binary classification with the WSI-level label available only. The 500 WSIs contain equal number of primary solid tumor images and solid tissue normal images. We randomly chose 400 WSIs (with 200 WSIs in each class) as the training set, and the rest as the validation set. For fair comparisons, we reproduced all competing methods using the same dataset split.
	
	The embedding matrix construction process is similar to the one used for Camelyon16. The WSI slide size is considerably smaller than the ones in Camelyon16, with an average number of 4872 patches. Therefore, we drawn only 100 patches from each of the 50 randomly selected WSIs in the positive training set, negative training set, positive validation set, and negative validation set respectively. 
	
	\begin{figure}[t]
    \centering
    \begin{subfigure}{.33\linewidth}
      \centering
      \includegraphics[width=1.0\linewidth]{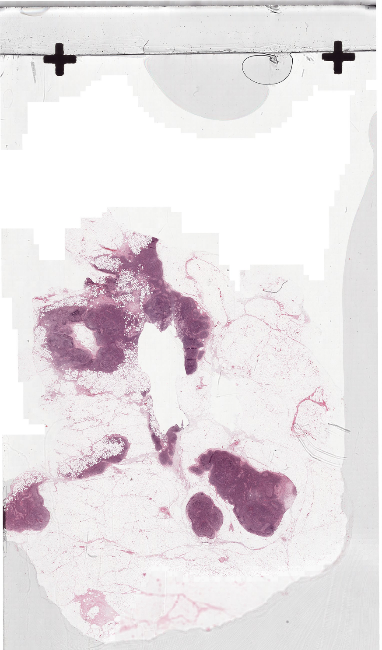}
      \caption{}
      \label{fig:sub1}
    \end{subfigure}%
    \begin{subfigure}{.33\linewidth}
      \centering
      \includegraphics[width=1.0\linewidth]{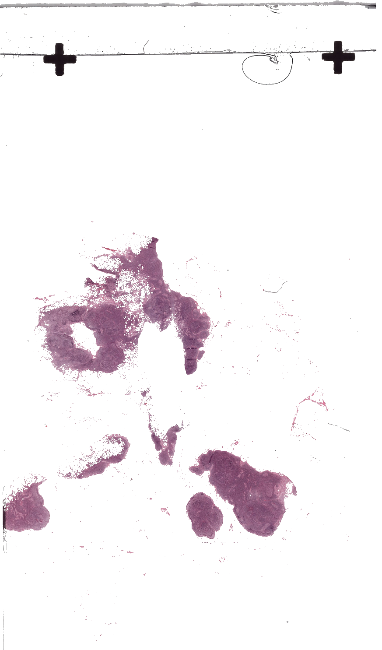}
      \caption{}
      \label{fig:sub2}
    \end{subfigure}%
    \begin{subfigure}{.33\linewidth}
      \centering
      \includegraphics[width=1.0\linewidth]{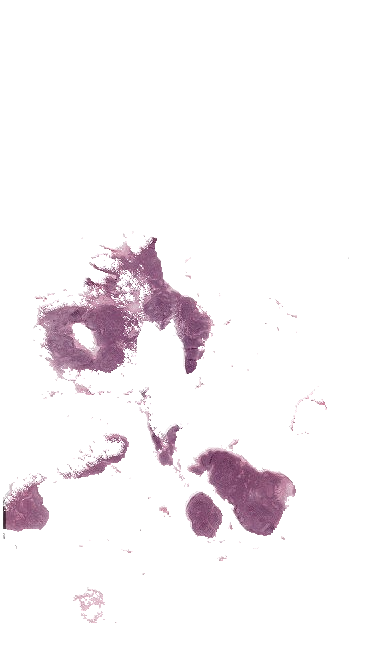}
      \caption{}
      \label{fig:sub3}
    \end{subfigure}
    \caption{\textbf{Pre-processing approaches.} (a) Raw WSI input. (b) Pre-processed by Otsu algorithm \cite{otsu1979threshold}. (c) Pre-processed by Run-length Encoding (RLE) \cite{robinson1967results} sorting and threshold (ours). (Background colours are reverted for better visualization.)}
    \label{fig:prep}
    \end{figure}
	
	\textcolor{black}{\paragraph{BCNB}
	Early Breast Cancer Core-Needle Biopsy WSI (BCNB) dataset includes core-needle biopsy whole slide images (WSIs) of early breast cancer patients and the corresponding clinical data. There are WSIs collected from a total of 1058 patients with only one WSI sampled from each patient. The entire dataset is officially \cite{bcnb} split into a training set, validation set, and testing set consisting of 630, 210, and 218 WSIs respectively. The WSIs are annotated from multiple perspectives including age, tumor size, tumor type, ER, PR, HER2, HER2 expression, histological grading, surgical, Ki67, molecular subtype, number of lymph node metastases, and the metastatic status of auxiliary lymph node (ALN). Following the official guideline \cite{bcnb}, we benchmarked our DSNet on this dataset for weakly supervised ALN classification with N0 as the negative class and N($+$) as the positive class. Unlike metastasis detection as in Camelyon16 and TCGA-LUSC, classifying the metastatic status of auxiliary lymph node in such a large dataset is much more challenging. The embedding matrix and thumbnail matrix construction processes are identical to the ones used for Camelyon16.}
	
	\subsection{Evaluation Metrics}
	Various metrics are used to evaluate the performance in the experiments: classification accuracy, precision, recall, and $\text{F}_{1}$-score. In addition, Receiver Operating Characteristic (ROC) curve and Area Under ROC (AUC) are also evaluated for comparing the performance of different methods.
	
	% Firstly, our method is evaluated on the dataset from Camelyon16 \cite{bejnordi2017diagnostic} challenge whose objective lies at the correct detection of cancer metastasis at WSI-level. The dataset consists of 400 sentinel lymph node hematoxylin and eosin (H\&E) WSIs from breast cancer patients. The 400 WSIs are split into 110 as positive training set (metastasis), 160 as negative training set (normal) and 130 comprised of 50 positive samples and 80 negative samples as testing set. Subsequently, we conduct experiments on the dataset from MICCAI15 WSI classification challenge with only 32 WSIs given to verify the generalization ability of our method with insufficient training data and by using basic data augmentation techniques only. The MICCAI15 dataset contains 16 WSIs for each of the two most common sub-types of lower grade gliomas brain cancer \cite{forst2014low}: astrocytoma (positive class) and  oligodendroglioma (negative class) respectively with no particular testing set. Different metrics are used to evaluate the performance in the experiments: classification accuracy, precision, recall and $\text{F}_{1}$-score. In addition, Receiver Operating Characteristic (ROC) curve and Area Under ROC (AUC) are also evaluated for comparing the performance of different methods.
	
		\begin{table}[t]
		\begin{center}
			\caption{AUC scores of fully-supervised, patch-wise, image-wise weakly-supervised methods over the validation sets of Camelyon16 (Cam16) and TCGA-LUSC (TCGA). `-' denotes unavailable results. Classifier parameters are reported as well.}
			\label{tab:1}
			\resizebox{\linewidth}{!}{\begin{tabular}{c |c c c c}
				\toprule
				%\Xhline{3\arrayrulewidth}
				Supervision & Methods & Cam16 & TCGA & \#Params \\
				\hline
				\hline
				%\multicolumn{3}{c}{\textit{Fully-supervised (upper bound)}}\\
				%\hline
				Fully & Wang \textit{et al.}. \cite{wang2016deep} & 0.925 & - & - \\
				\hline
				%\multicolumn{3}{c}{\textit{Patch-wise weakly-supervised}}\\
				%\hline
				\multirow{3}{1.4cm}{\centering Patch-wise\\Weakly}
				 & Hou \textit{et al.}. \cite{hou2015efficient}  & 0.666 & 0.973 & 25.6 M\\
				 & Chen \textit{et al.}. \cite{chen2019rectified} &0.643 & 0.975 & 25.6 M\\
				& VAE-GAN \cite{zhao2020predicting} &0.695 & 0.976 & 32.5 M\\
				\hline
				%\multicolumn{3}{c}{\textit{Image-wise weakly-supervised}}\\
				%\hline
				\multirow{3}{1.4cm}{\centering Image-wise\\Weakly}
				& Naive& 0.524 & 0.956 & 25.6 M\\\
				 & Tellez \textit{et al.}. \cite{tellez2019neural} &0.717 & 0.976 & 25.6 M\\
				&\textbf{DSNet (ours)} & \textbf{0.760} & \textbf{0.986} & \textbf{1.1} M\\
				\bottomrule
				%\Xhline{3\arrayrulewidth}
			\end{tabular}}
		\end{center}
\end{table}

\begin{table}[t]
		\begin{center}
			\caption{\textcolor{black}{AUC scores of the baseline image-level method, and our DSNet over the validation set and testing set of BCNB.}}
			\label{tab:bcnb}
			\resizebox{\linewidth}{!}{\begin{tabular}{c |c c c c}
				\toprule
				%\Xhline{3\arrayrulewidth}
				\textcolor{black}{\ Supervision\ } & \textcolor{black}{Methods} & \textcolor{black}{Val.} & \textcolor{black}{Test.} & \textcolor{black}{\#Params} \\
				\hline
				\hline
				\textcolor{black}{Patch-wise} & \multirow{2}{1.6cm}{\centering \textcolor{black}{Xu \textit{et al.} \cite{bcnb}}} & \multirow{2}{0.8cm}{\centering \textcolor{black}{0.808}} & \multirow{2}{0.8cm}{\centering \textcolor{black}{0.816}} & 
				\multirow{2}{0.8cm}{\centering \textcolor{black}{4.1 M}} \\
				\textcolor{black}{Weakly} &  \\
				\hline
				\textcolor{black}{Image-wise}
				& \textcolor{black}{Naive}& \textcolor{black}{0.573} & \textcolor{black}{0.554} & \textcolor{black}{25.6 M}\\\
				\textcolor{black}{Weakly} &\textcolor{black}{DSNet (ours)} & \textcolor{black}{0.797} & \textcolor{black}{0.803} & \textcolor{black}{\textbf{1.1} M}\\
				% 0.808, 0.816
				\bottomrule
				%\Xhline{3\arrayrulewidth}
			\end{tabular}}
		\end{center}
\end{table}
	
\begin{table*}[t]
		\begin{center}
			\caption{Ablation study on the Camelyon16 validation set and TCGA-LUSC validation set. We denote \textbf{Sa} as the Stream-wise attention, \textbf{Ca} the Channel-wise attention.}
			\label{tab:ablation}
			\resizebox{\linewidth}{!}{\begin{tabular}{c|c c c c c|c c c c c}
					\toprule
					%\Xhline{3\arrayrulewidth}
					& \multicolumn{5}{c|}{Camelyon16} & \multicolumn{5}{c}{TCGA-LUSC}\\\cline{2-6} \cline{7-11}
					Methods & Accuracy & AUC & Precision & Recall & \multicolumn{1}{c|}{$\text{F}_1$-score} & \multicolumn{1}{c}{Accuracy} & AUC & Precision & Recall & $\text{F}_1$-score \\
					\hline
					\hline
					%\Xhline{2\arrayrulewidth}
					w/o sparsity & 67.8\% & 0.635 & 0.209 & \textbf{0.750} & 0.327 & 89.0\% & 0.977 & 0.860 & 0.915 & 0.887\\
					%\rowcolor{black}
					\textcolor{black}{w/o embedding stream} & \textcolor{black}{57.7\%} & \textcolor{black}{0.536} & \textcolor{black}{0.439} & \textcolor{black}{0.360} & \textcolor{black}{0.396} & \textcolor{black}{83.0\%} & \textcolor{black}{0.959} & \textcolor{black}{0.867} & \textcolor{black}{0.780} & \textcolor{black}{0.821} \\
					w/o thumbnail stream & 60.9\% & 0.686 & 0.209 & 0.450 & 0.286& 89.0\% & 0.965 & 0.950 & 0.780 & 0.857 \\
					w/o multi-scale block & 59.1\% & 0.670 & 0.674 & 0.467 & 0.552 & 85.0\% & 0.974 & 0.780 & 0.907 & 0.839 \\
					w/o concurrent bottlenecks & 66.0\% & 0.656 & 0.395 & 0.515 & 0.447 & 89.0\% & 0.968 & 0.840 & 0.933 & 0.884 \\
					w/o spatial bottleneck & 60.0\% & 0.696 & \textbf{0.721} & 0.477 & 0.574 & 82.0\% & 0.964 & 0.680 & 0.944 & 0.791 \\
					w/o \textbf{Sa} w/o \textbf{Ca} & 67.0\% & 0.714 & 0.395 & 0.586 & 0.472 & 90.0\% & 0.975 & 0.880 & 0.916 & 0.898 \\
					w/o \textbf{Sa} w. \textbf{Ca} & 63.5\% & 0.695 & 0.349 & 0.517 & 0.417 & 85.0\% & 0.975 & 0.780 & 0.907 & 0.839 \\
					w. \textbf{Sa} w/o \textbf{Ca} & 65.3\% & 0.709 & 0.349 & 0.556 & 0.429 & 84.0\% & 0.975 & 0.780 & \textbf{0.886} & 0.830 \\
					
					%Courtiol \textit{et al.} \cite{DBLP:journals/corr/abs-1802-02212} & 0\% & 0 & 0 & 0 & 0 & 0\% & 0 & 0 & 0 & 0 \\
					\hline
					\textbf{Full DSNet} & \textbf{69.6\%} & \textbf{0.760} & 0.558 & 0.600 & \textbf{0.578}& \textbf{93.0\%} & \textbf{0.986} & \textbf{1.000} & 0.877 & \textbf{0.935}\\
					
					%\Xhline{2\arrayrulewidth}
				% 	DSNet + GAP \cite{DBLP:journals/corr/LinCY13} & 55.0\% & 0.613 & 0.632 & 0.670 & 0.650 & 84.4\% & 0.935 & 0.789 & 0.938 & 0.857\\
				% 	DSNet + GSP & \textbf{68.7\%} & 0.728 & 0.705 & 0.861 & \textbf{0.775} & \textbf{93.8\%} & 0.943 & \textbf{1.000} & 0.875 & 0.933 \\ 
				% 	DSNet + TOP \cite{durand2016weldon,DBLP:journals/corr/abs-1802-02212} & 67.0\% & \textbf{0.729} & \textbf{0.743} & 0.722 & 0.732 & 93.8\% & \textbf{0.961} & 0.938 & \textbf{0.938} & \textbf{0.938} \\
					
					\bottomrule
					%\Xhline{3\arrayrulewidth}
			\end{tabular}}
		\end{center}
	\end{table*}
	
	\subsection{Image Pre-processing} \label{pre}

    \textcolor{black}{Before preparing all of the datasets}, we applied extra WSI pre-processing procedures to exclude tissue fluid and blood to enhance the accuracy of our method and to reduce computation costs. We then crop the tightest bounding box on the foreground pixels to focus on the slide regions only. Note that we did not stain normalize the WSIs in our experiments, as we found that such pre-processing empirically led to poorer performances.
    
    WSI pre-processed results with different approaches are visualized in Figure \ref{fig:prep}. The commonly used Otsu \cite{otsu1979threshold} algorithm cannot effectively remove the ink marks on the WSI slides. By applying the Run-Length Encoding (RLE) \cite{robinson1967results} on the average pixel intensity of patches, a threshold can be easily decided between the large encoding patches (rich pixel variations, i.e. patches with cells) and the small encoding patches (i.e. patches with tissues/inks).
    	
	\subsection{Implementations} \label{imp}
	
	\textcolor{black}{We implemented all experiments on a NVIDIA RTX 2080 GPU using the Keras framework \cite{chollet2015keras}.}
	
	\textcolor{black}{\paragraph{Hyper-parameter settings} There are three main hyper-parameters considered in our framework: embedding matrix channel number $\text{C}$, patch size $\text{S}$, and the greatest thumbnail downsampling rate $\text{K}$. $\text{C}$ = 128 and $\text{S}$ = 256 are directly borrowed from Tellez \textit{etal}. \cite{tellez2019neural}. For the thumbnail image downsampling rate K, we followed a simple intuition that sets K = 128 to be half of the patch size. In this way, the thumbnail image will contain at least four times more pixels than the embedding matrix and provide much more abundant regional information. All of these three hyper-parameters are adopted as default without any explicit tuning. Moreover, in our early trials, we found that tuning these hyperparameters contributed little to the overall framework performance. For better consistency and generalization ability, we used the most intuitive hyperparameters and kept them unchanged across all our experiments.}

	\paragraph{S-CAE implementation details} In all experiments related to S-CAE, we use SGD optimizer with Nesterov momentum to minimize the mean squared error (MSE) between inputs and outputs. The initial learning rate is set to 0.03, weight decay to 0.00001, momentum to 0.8, and batch size to 8. We train the S-CAE for 6 epochs for each of the selected WSIs on both datasets. All patches are extracted at the highest magnification level. Foreground and background pooled embeddings are concatenated with a split of 3:1. 

	Training patches are extracted in size 256 $\times$ 256 and then resized to 112 $\times$ 112 to fit the S-CAE. 600 patches are drawn from each of the 20 randomly selected WSIs in the positive training set, negative training set, and validation set respectively. In order to better encode the patches with a rich distribution of nuclei, we ensure at least 90\% of the training patches are drawn with all three channels' intensities being less than 70\% of the maximum intensity, which represents samples from rich nucleus regions. 
	
	\paragraph{DSNet implementation details} Our DSNet is constructed as described in Sec. \ref{DCR}, which is supervised by a single cross-entropy loss. Leaky ReLU is used as the activation function for convolutions. 
	
	For training configurations, the loss of DSNet is minimized through the gradient descent algorithm with AdamW optimizer \cite{loshchilov2018decoupled}. The initial learning rate is set to $1e^{-4}$ with warm up from $1e^{-6}$ and following a step decay by a factor of 5 at every 30 epochs. We set weight decay to $1e^{-5}$. All models are trained for 200 epochs from scratch with early stopping at the epoch with the highest AUC score. We augment all training data by standard image flipping, rotating, and random cropping with discarding at most 7 pixels along the height or width.
	
	\subsection{Benchmark on Camelyon16}
	
	Our method is first evaluated on the dataset from Camelyon16 \cite{bejnordi2017diagnostic} challenge for detection of cancer metastasis at WSI-level. 
	
	We compared with the fully-supervised method (Camelyon16 challenge winner) \cite{wang2016deep}, state-of-the-art weakly-supervised methods through patch-wise approaches \cite{hou2016patch, chen2019rectified} and image-wise approach \cite{tellez2019neural}. We also conducted experiments on a naive image-wise approach, by using ResNet-50 to classify the down-sampled WSI thumbnail image directly. The fully-supervised method requires ground truth annotations to indicate patch-level labels during their training phase, while only the WSI-level label is used in our method during the whole training process.
	
	 We reproduced all weakly-supervised methods \cite{hou2016patch, chen2019rectified, tellez2019neural,zhao2020predicting} for fair comparison \footnote{\cite{zhao2020predicting} originally relies on extra annotations from the WSI experts, while its unavailable in our weakly-supervised settings.} except for the fully-supervised method \cite{wang2016deep} on Camelyon16, which we report the result from their paper directly. Note that all weakly-supervised methods claimed using standard networks for classification, and we keep the training configurations and classifier networks (ResNet-50 \cite{he2016deep}) consistent across all weakly-supervised approaches for fair comparisons.

	Table \ref{tab:1} reports the performances of the methods in terms of AUC scores and total number of network parameters \footnote{For image-wise approaches, patch encoders require extra parameters, which are 1.8M for \cite{tellez2019neural} and 2.4M for our S-CAE.}. It can be seen that the naive image-level weakly-supervised approach performs the worst, as large amount of visual information is lost when downsizing the WSIs. On the contrary, with the help of a previous encoding of raw WSIs, image-wise approaches are able to maintain useful diagnostic clues to a great extent and achieve superior performances than the patch-wise approaches. Particularly, our DSNet, with an advanced encoder and a dedicated classifier, eventually achieves an AUC of 0.760 that outperforms recent state-of-the-art weakly-supervised counterparts by a safe margin. Noteworthy, our proposed DSNet requires only 1.1 M trainable parameters, which is around 4\% of the standard ResNet-50 classifier. The ROC curves of our method along with the state-of-the-art weakly supervised methods are plotted in Figure \ref{fig:roc1}.

	\subsection{Benchmark on TCGA-LUSC}
	
	% \footnote{Our dataset split will be released for facilitating any further research.}
	
	Evaluation results are presented in Table \ref{tab:1}, the component study results are presented in Table \ref{tab:ablation}, and the comparison ROC curves are plotted in Figure \ref{fig:tcga_roc}. Compared to other weakly-supervised counterparts, our method achieves the best performance in terms of the AUC scores. 
	
	\subsection{Benchmark on BCNB}
	
	We further evaluate our DSNet and the naive image-wise baseline on a larger dataset for classifying the metastatic status of auxiliary lymph node. The comparison results of the method proposed in \cite{bcnb} and the naive image-wise baseline are reported in Table \ref{tab:bcnb}. With the minimum network size, our DSNet significantly surpasses the naive image-wise counterpart and achieves on par results compared to the state-of-the-art method on this dataset \cite{bcnb}.
	
	Although the patch-wise state-of-the-art method proposed in \cite{bcnb} yields slightly better results than ours, their method \textit{(i)} used extra data (ImageNet) for model pre-training, while all of our models were trained using in-distribution data only; \textit{(ii)} was built with nearly 4 times more trainable parameters.
	
	%\footnote{We will release our TCGA dataset split for public benchmark.}
	\section{Analysis}
	In this section, we provide in-depth analysis and additional ablation studies for a comprehensive understanding of the proposed method. Unless explicitly specified, the extensive experiments were conducted on the Camelyon16 dataset, with the identical training protocols introduced in Sec. \ref{imp}.
	
		\begin{table}[t]
		\begin{center}
			\caption{Comparison of using different levels of thumbnail images on the basis of downsampling ratio \text{K}=128.}
			% \textcolor{black}{$\alpha$ is the scalar multiplier to \text{K}}.
			\label{tab:level}
			\resizebox{\linewidth}{!}{\begin{tabular}{l c c c c c}
					\toprule
					%\Xhline{3\arrayrulewidth}
					\textcolor{black}{$\frac{\text{K}}{2^\alpha}$} & Accuracy & AUC & Precision & Recall & Comp ratio  \\
					\hline
					\hline
					%\Xhline{2\arrayrulewidth}
				    $\alpha$=[0] & 67.8\% & 0.736 & 0.558 & 0.571 & \textcolor{black}{\textbf{1404}} \\%\textbf{1031} \\
					$\alpha$=[0,1] & 67.0\% & 0.759 & \textbf{0.604} & 0.553 & \textcolor{black}{1046}\\%776 \\
					$\alpha$=[0,1,2] & \textbf{69.6\%} & \textbf{0.760} & 0.558 & 0.600 & \textcolor{black}{517}\\%622 \\
					$\alpha$=[0,1,2,3] & 68.7\% & 0.755 & 0.535 & 0.590 & \textcolor{black}{171}\\%519 \\
					$\alpha$=[0,1,2,3,4] & 67.0\% & 0.737 & 0.233 & \textbf{0.714} & \textcolor{black}{56}\\%445 \\
					\bottomrule
					%\Xhline{3\arrayrulewidth}
			\end{tabular}}
		\end{center}
	\end{table}
	
	\subsection{Complexity}
		
	\paragraph{Space complexity} The hardware-unaffordable image size is the primary challenge to WSI analysis. Our method transforms the gigabyte original WSI to a latent space in a unsupervised style for reducing space complexity. Given the thumbnail downsampling factor $\text{K}$ and the number of level $\text{L}$, the \textit{thumbnail matrix} achieves a compression ratio of $\frac{\text{K}^2}{\text{L}}$. With the patch size S, the \textit{embedding matrix} is able to achieve a compression ratio of $\frac{\text{3S}^2}{\text{C}}$. \textcolor{black}{We then represent the original WSI by the two matrices, with the final compression ratio (w.r.t. \# pixels) of $\text{3S}^{2}\text{K}^{2} / (\text{C}\text{K}^2 + (\sum_{l=0}^{L-1}2^{2l}3)\text{S}^2)$. With our adopted setting (S=256, C=128, K=128, L=3), a compression ratio of 517 can be achieved}.
	
		\begin{table}[t]
		\begin{center}
			\caption{Comparison of different patch encoding strategies.}
			\label{table:encoding}
			\resizebox{\linewidth}{!}{\begin{tabular}{c c c c c c}
					\toprule
					%\Xhline{3\arrayrulewidth}
					Encoding & Accuracy & AUC & Precision & Recall & $\text{F}_1$-score \\
					\hline
					\hline
					Foreground & 57.4\%&0.650&0.302&0.406&0.347\\
				    Background & 58.3\%&0.585& 0.233&0.400&0.294 \\
				    Mixed & 67.0\%& 0.696 & 0.233&\textbf{0.667}&0.345\\
					Separated & \textbf{69.6\%} & \textbf{0.760} &\textbf{0.558} & 0.600 & \textbf{0.578} \\
					\bottomrule
					%\Xhline{3\arrayrulewidth}
			\end{tabular}}
		\end{center}
	\end{table}
	
		\begin{figure*}[t]
		\minipage{0.32\textwidth}
		\includegraphics[width=\linewidth]{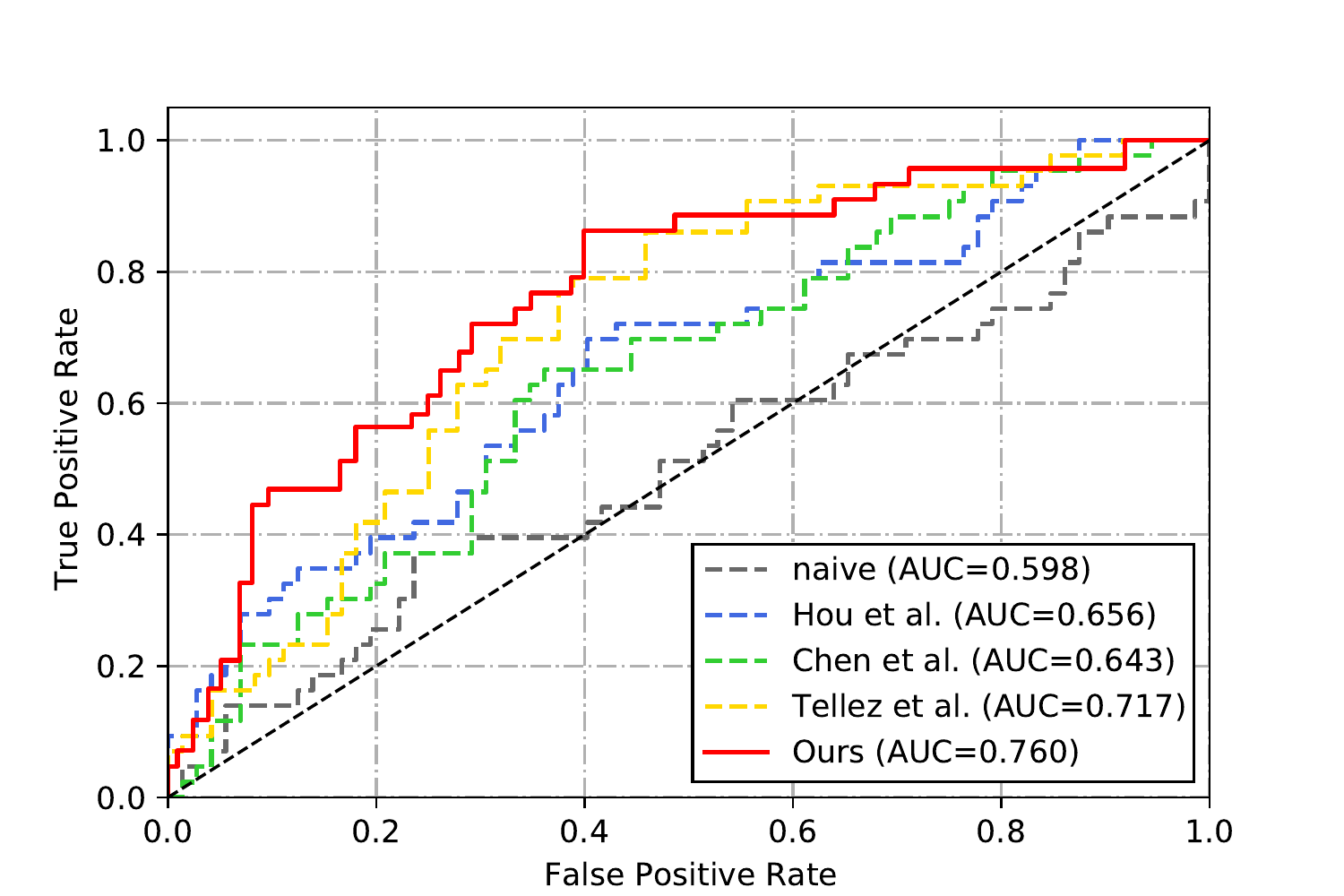}
		\caption{ROC curves of comparisons to the state-of-the-art weakly-supervised methods on the Camelyon16 validation set.}\label{fig:roc1}
		\endminipage\hfill
		\minipage{0.32\textwidth}%
		\includegraphics[width=\linewidth]{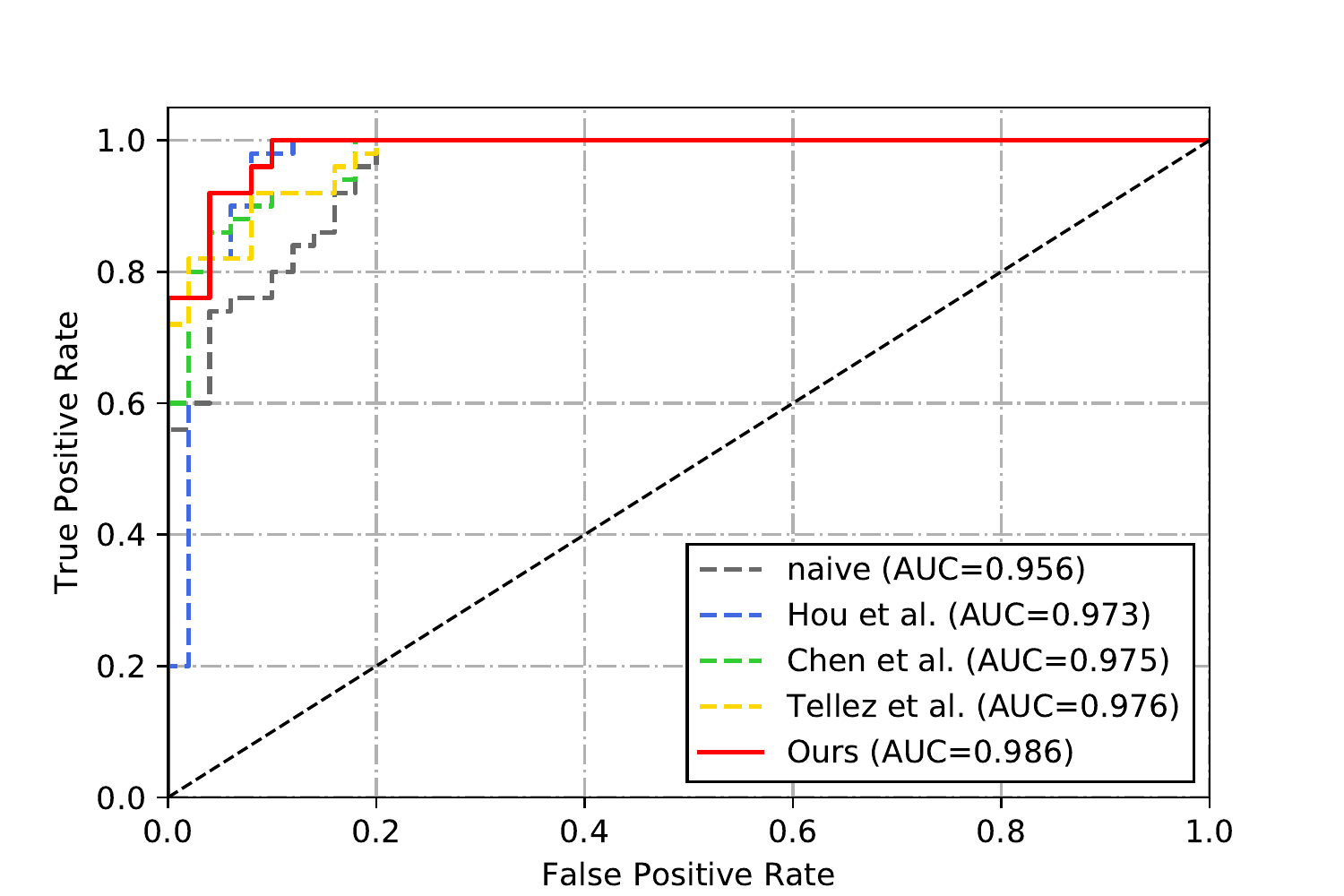}
		\caption{ROC curves of comparisons to the state-of-the-art weakly-supervised methods on the TCGA-LUSC validation set.}\label{fig:tcga_roc}
		\endminipage\hfill
		\minipage{0.32\textwidth}
% 		\includegraphics[width=\linewidth]{image/bar.pdf}
% 		\caption{On the Camelyon16 validation set, an average stream-wise attention score of 0.539 is applied on local stream.}\label{fig:bar}
		\includegraphics[width=\linewidth]{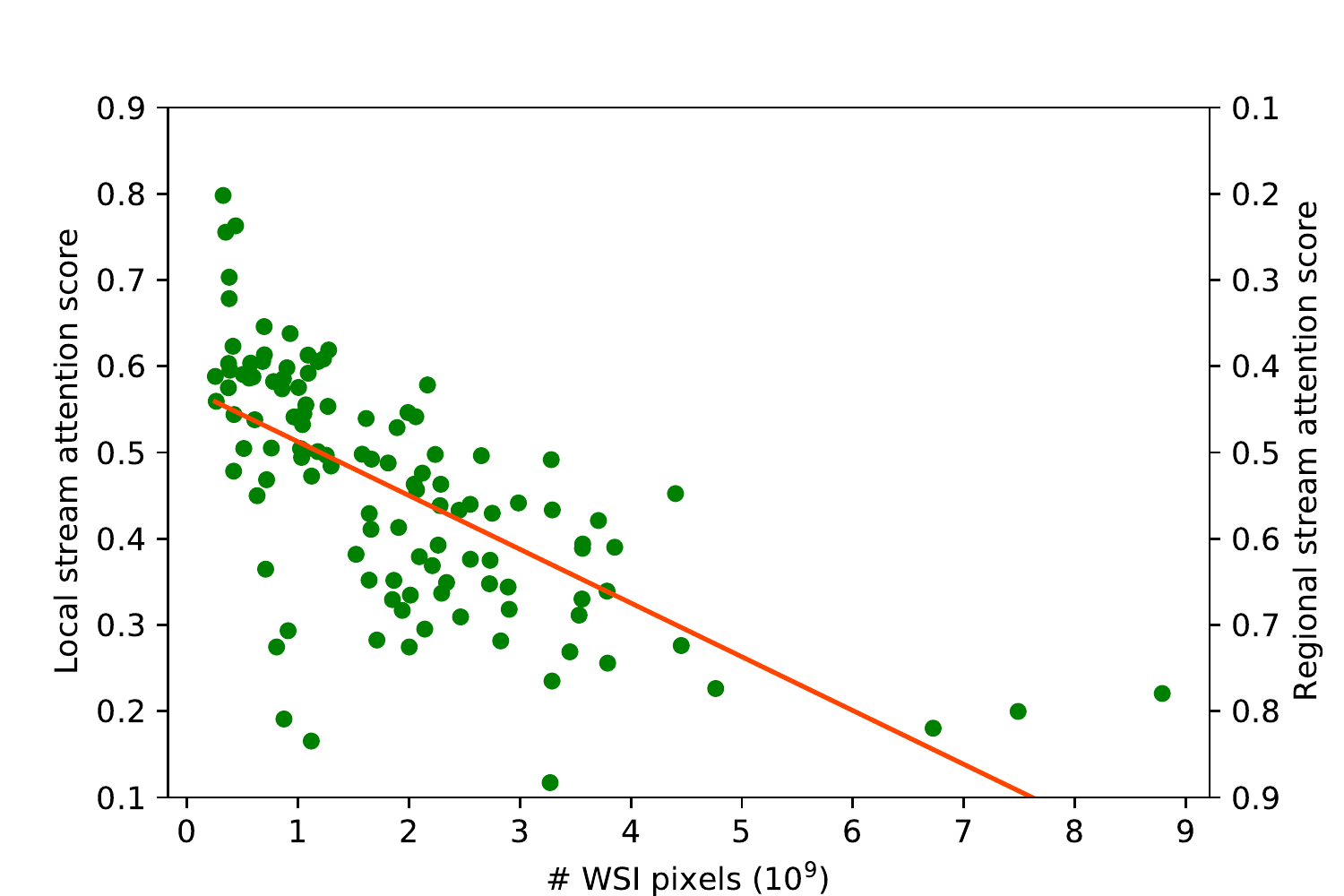}
		\caption{\textcolor{black}{Stream-wise attention score \textit{v.s.} \# WSI pixels on the Camelyon16 validation set. A linear fit on the attributes (red line) is shown.}}\label{fig:bar}
		\endminipage
	\end{figure*}
	
	\begin{figure*}[t]
		\begin{center}
			%\fbox{\rule{0pt}{2in} \rule{.9\linewidth}{0pt}}
			\includegraphics[width=1.0\linewidth]{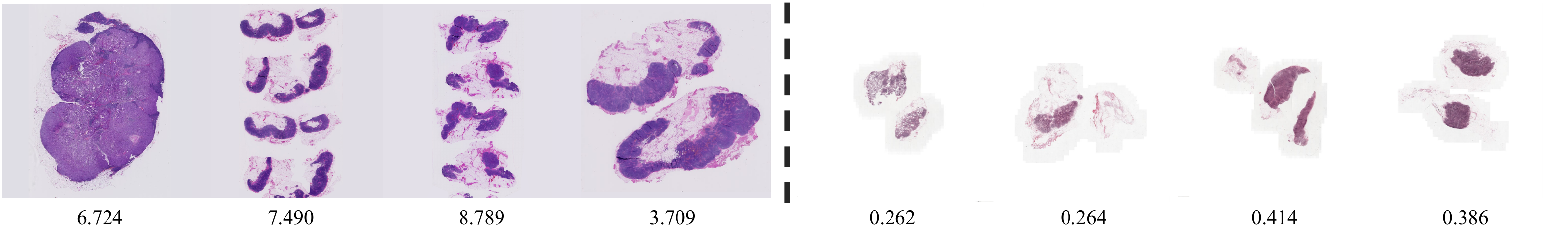}
		\end{center}
		\caption{\textbf{Visualization of local and regional attentive WSI examples}. Slide areas ($\times$10$^9$ pixel) are shown in this figure. The WSIs on the left are paid more attention on regional information ($<$30\% attention scores are assigned to local stream) and the right ones are paid more attention on local information ($>$80\% attention scores are assigned to local stream).}
		\label{fig:instance}
	\end{figure*}
	
	\paragraph{Time complexity} \textcolor{black}{Time complexity is another essential consideration in our method development. The depth-wise design \cite{howard2017mobilenets} was adopted as our basic convolutional operator to speed up network inference and lower the computation burden. When the thumbnail/embedding matrices are available, DSNet inference on a WSI with $10^9$ pixels takes about \textit{44.1 ms} on a NVIDIA RTX 2080 GPU and \textit{273.2 ms} on an Intel Core i7 CPU. Moreover, our entire pipeline, which includes: \textit{(i)} WSI file loading from a hard disk; \textit{(ii)} image preprocessing (patch dividing + patch sifting + matrices formation); and \textit{(iii)} DSNet inference, processes a WSI with $10^9$ pixels in \textit{126.1 s} on a CPU. The overall runtime can be reduced to \textit{63.6 s} if a GPU is available. All running times were measured over an average of 10 WSIs at $40\times$ magnification level.}
	
	\subsection{Ablation Studies}
    We examine the impact of individual components by removing each of them from the network. To assess the necessity of considering sparsity issues in WSIs, experiments are carried out by using dense operators only. \textcolor{black}{Then, the entire embedding stream or thumbnail stream is removed from DSNet to verify whether the local or regional information is useful to the metastasis diagnosis. Subsequently, we examined the effectiveness of the multi-scale block by simply replacing it with $3\times3$ convolution only, and aligned the total network parameters.} We further assess the impact of the concurrent bottleneck block and adaptive aggregation block separately. The ablation results are shown in Table \ref{tab:ablation}. Without considering the sparsity issue, the network can hardly detect metastasis and tends to make predictions toward the negative class. The designed concurrent bottleneck block also yields a significant impact on our network by raising the AUC score for more than 10\%. Surprisingly, we found that even without adaptive aggregation, our method can still achieve a competitive result at 0.714 AUC. \textcolor{black}{However, the performance of DSNet could be harmed when only one attention operator is used. Our hypothesis is that the features suppressed in one perspective (e.g. channel-wise significance) could be more indicative in another perspective (e.g. stream-wise significance) \cite{woo2018cbam}. Therefore, a single measurement of the feature significance in our adaptive aggregation module could degenerate feature representation.} When integrating all the components into our network, the full DSNet reaches the state-of-the-art performance at 69.6\% classification accuracy and 0.760 AUC.
	
	%With \textbf{DA} block added, compared with the baseline method, our DSNet with GSP improves accuracy, AUC, precision, recall and $\text{F}_1$-score by 17.4\%, 0.139, 0.075, 0.319 and 0.193 respectively. With the top instances only, our DSNet achieves an even better AUC score of 0.729.

	\subsection{Level of Thumbnail Images} 
	The thumbnail matrix of a WSI is constructed as a stack of thumbnail images captured at different magnified levels. We study the impact of using thumbnails at different magnifications as input to our DSNet. In Table \ref{tab:level}, we achieve 0.736 AUC by only using the image at the highest magnification on a downsampling basis of K=128, which already surpasses the image-wise state-of-the-art counterpart \cite{tellez2019neural}. Compromising on the compression ratio, our method reaches the highest AUC score by using thumbnail images from three different levels.
	
			\begin{figure}[t]
	\begin{center}
		\includegraphics[width=1.0\linewidth]{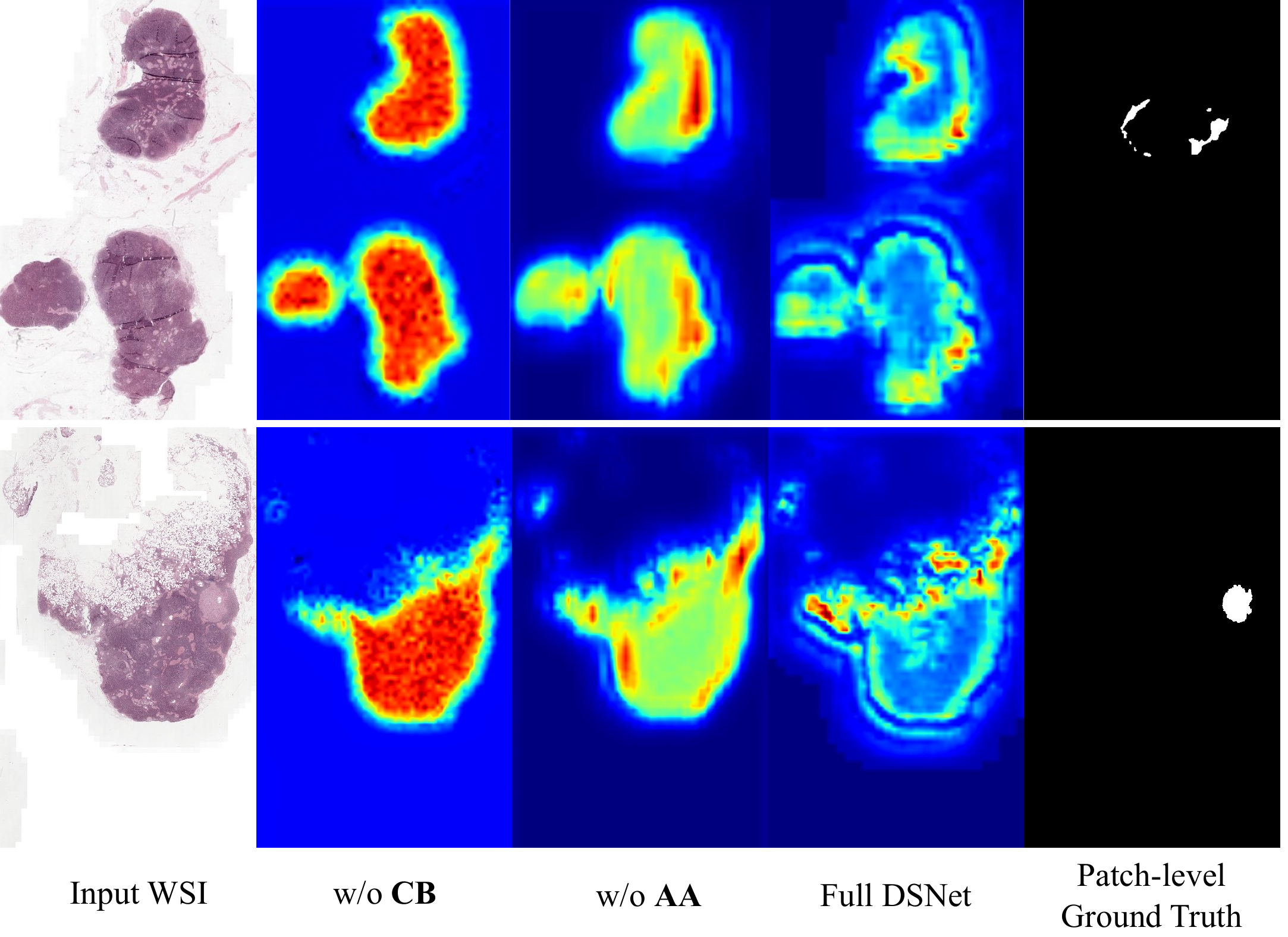}
	\end{center}
	\caption{\textbf{Grad-CAM visualizations of our \textit{weakly-supervised} DSNet}. \textcolor{black}{$\mathbf{AA}$ denotes our Adaptive Aggregation operator and $\mathbf{CB}$ denotes our Concurrent Bottleneck design.}}
	\label{fig:gradcam}
\end{figure}

	\subsection{Patch Encoding Strategy}
	Here, we evaluate different patch encoding strategies by using foreground encoding only, background encoding only, and the mix encoding without distinguishing foreground or background. Table \ref{table:encoding} shows that using both separated foreground and background encodings yields the best performance at nearly all metrics. Similar to \cite{tellez2019neural}, by using a mixed encoding, we can also obtain a good result.
	
	\subsection{Stream Importance} 
	We then analyze the importance of the two streams regrading of the learned stream-wise attention scores. \textcolor{black}{The stream-wise attention score distribution on the Camelyon16 validation set with respect to the WSI size is shown in Figure \ref{fig:bar}. As the figure illustrates, our DSNet is more favorable to assign higher attention score to the local stream (embedding matrix stream) when the size of input WSI is small and to the regional stream (thumbnail matrix stream) when the size is large. To understand whether our DSNet can demonstrate certain correlations between WSI size and the attention score, we first calculate an absolute Pearsons correlation coefficient of \textit{0.656} and fit a linear regression model on the two attributes (Figure \ref{fig:bar}). We further visualize some instances with most local attention scores and most regional attention scores in Figure \ref{fig:instance}. From the Pearsons coefficient, the fitted linear model, and individual instances, we found that our DSNet indeed has the ability to assign different stream-wise attention scores regarding of WSI slide areas. Interestingly, pathologists are supposed to pay more attention to regional clues when diagnosing larger WSIs and focus more on local details for smaller WSIs, which is consistent with our findings}.
	
% 	pathologists are supposed to pay more attention to regional information than local information when diagnosing WSIs with billions of pixels, which is consistent with what our method reveals. 

	\subsection{Grad-CAM Visualization} 
	Grad-CAM \cite{selvaraju2017grad} is a commonly used algorithm to visualize the region of network interest toward the prediction of a target class. We use Grad-CAM to highlight the area that our DSNet determines as evidences to diagnose a WSI as containing metastasis. The gradients are collected at the last $1\times 1$ convolutional layer towards the positive class. In Figure \ref{fig:gradcam}, we present the comparison results by removing different components from the network to inspect the changes of network focus. When removing the concurrent bottleneck, the base network can hardly discriminate local semantics, results in paying attention to all valid regions. Without patch-level labels, our weakly-supervised method is also able to indicate reasonable regions that may show potential cancer metastasis. 
	
	%More Grad-CAM visualization results are shown in the supplemental materials

	\section{Conclusion} \label{con}
	
	We propose a novel framework for classifying whole slide images in a weakly-supervised manner to alleviate the burden of expert annotation that is required in fully supervised approaches. Our method consists of two sequential stages, which utilize an efficient neural network to first roughly skim through a WSI thumbnail and then zoom into regions of interest for more descriptive details. Specifically, our framework transforms the raw WSI in an unsupervised manner into a local embedding matrix and a regional thumbnail matrix, and then employs the proposed DSNet to adaptively integrate the transformed matrices to infer the WSI-level label through two separated streams consisting of stacks of efficient yet effective building blocks. We demonstrate that our image-level weakly-supervised approach surpasses recent state-of-the-art methods on three large-scale public benchmark datasets with the least network parameters.
	
	Our proposed weakly-supervised approach processes gigapixel WSIs at their compact representations with minimum computational costs and hardware burdens. With such highly efficient two stream design, regional guidances implied in thumbnail images could be useful in all kinds of WSI analysis applications. Therefore, we believe the same framework can be easily extended to many other tasks such as metastasis segmentation and anomaly detection in future works. 

{\small
\bibliographystyle{style2}
\bibliography{IEEEabrv, egbib}
}


% \bibitem{b1} G. O. Young, ``Synthetic structure of industrial plastics,'' in \emph{Plastics,} 2\textsuperscript{nd} ed., vol. 3, J. Peters, Ed. New York, NY, USA: McGraw-Hill, 1964, pp. 15--64.

% \end{thebibliography}

\end{document}